%% file: main.tex
\documentclass[10pt,twocolumn,letterpaper]{article}
\usepackage{multirow}   % enables \multirow
\usepackage{algorithmic}
\usepackage{booktabs}   % enables \toprule, \midrule, \bottomrule
\usepackage{array}      % better column types (e.g., p{...})
\usepackage{booktabs}       % for professional looking tables
\usepackage[table]{xcolor}  % for row coloring
\usepackage{multirow}       % if you want to merge rows later
\usepackage{amsmath,amssymb,booktabs,multirow,graphicx,algorithm,algorithmic,xcolor}
\usepackage{comment}
\usepackage{graphicx}
\usepackage{arydshln}
\usepackage{tikz}
\usetikzlibrary{arrows.meta,positioning,fit,shapes.misc,shapes.geometric,calc}
\usepackage{colortbl}
\usepackage{amssymb}
\usepackage{algorithmic}
\usepackage{algorithm}
\usepackage{cuted}     % <- provides the strip environment
\usepackage{capt-of}
\usepackage[table]{xcolor}
\usepackage{booktabs}
\usepackage[table]{xcolor}
\usepackage{multirow}
\usepackage{siunitx}
\usepackage{booktabs}
\usepackage[table]{xcolor}
\usepackage{setspace}
\usepackage{lipsum}
\usepackage{graphicx}
\usepackage{booktabs}
\usepackage{amsmath}
\usepackage{amssymb}
\usepackage{colortbl}
\usepackage{multirow}
\usepackage{xcolor}
\usepackage{subcaption}
\usepackage{pifont}
\newcommand{\cmark}{\textcolor{green!60!black}{\ding{51}}}
\newcommand{\xmark}{\textcolor{red!70!black}{\ding{55}}}
\usepackage[accsupp]{axessibility}

\usepackage{orcidlink}
\usepackage{soul}

\usepackage{cvpr}      % To produce the REVIEW version


\definecolor{cvprblue}{rgb}{0.21,0.49,0.74}
\usepackage[pagebackref,breaklinks,colorlinks,allcolors=cvprblue]{}
\usepackage{comment}
\usepackage{fontawesome5}

\title{\faMapMarked\ VideoAtlas: Navigating Long-Form Video in Logarithmic Compute}

\author{%
\makebox[\textwidth][c]{%
  \begin{tabular}{@{}c c c@{}}
  Mohamed Eltahir$^{1}$ &
  Ali Habibullah$^{1}$\footnotemark[1] &
  Yazan Alshoibi$^{1}$\footnotemark[1] \\
  Lama Ayash$^{1,2}$ &
  Tanveer Hussain$^{3}$\footnotemark[2] &
  Naeemullah Khan$^{1}$\footnotemark[3] 
  \end{tabular}%
}%
\\
$^{1}$ King Abdullah University of Science and Technology (KAUST), Thuwal, Saudi Arabia \\
$^{2}$ Department of Computer Science, King Khalid University (KKU), Abha, Saudi Arabia \\
$^{3}$ Department of Computer Science, Edge Hill University, Ormskirk, England \\
\texttt{\{mohamed.hamid, ali.habibullah, yazen.shaebi, lama.ayash\}@kaust.edu.sa} \\
\texttt{hussaint@edgehill.ac.uk, naeemullah.khan@kaust.edu.sa}}

\begin{document}

\addtolength{\topmargin}{-1.5cm}
\addtolength{\textheight}{1.8cm}

\maketitle

\begin{abstract}
Extending language models to video introduces two challenges: \emph{representation}, where existing methods rely on lossy approximations such as uniform sampling, and \emph{long-context}, where caption- or agent-based pipelines collapse video into text and lose visual fidelity. To overcome this, we introduce VideoAtlas, a task-agnostic environment to represent video as a hierarchical grid that is simultaneously lossless, navigable, scalable, caption- and preprocessing-free. An overview of the video is available at a glance, and any region can be recursively zoomed into, with the same visual representation used uniformly for the video, intermediate investigations, and the agent's memory, eliminating lossy text conversion end-to-end. This hierarchical structure ensures access depth grows only logarithmically with video length.
For long-context, Recursive Language Models (RLMs) recently offered a powerful solution for long text, but extending them to visual domain requires a structured environment to recurse into, which VideoAtlas provides. VideoAtlas as a Markov Decision Process unlocks Video-RLM: a parallel Master-Worker architecture where a Master coordinates global exploration while Workers concurrently drill into assigned regions to accumulate lossless visual evidence. 
We demonstrate three key findings: (1)~logarithmic compute growth with video duration, in contrast to the linear cost of baselines, further amplified by a 30-60\% multimodal cache hit rate arising from the grid's structural reuse. (2)~environment budgeting, where bounding the maximum exploration depth provides a principled compute-accuracy hyperparameter. (3)~emergent adaptive compute allocation that scales with question granularity. When scaling from 1-hour to 10-hour benchmarks, Video-RLM remains the most duration-robust method with minimal accuracy degradation while baselines degrade significantly, demonstrating that structured environment navigation is a viable and scalable paradigm for video understanding.
\end{abstract}

{
  \renewcommand{\thefootnote}{\fnsymbol{footnote}}
  \footnotetext[1]{Equal Contribution}
  \footnotetext[2]{Corresponding Author}
  \footnotetext[3]{Principal Investigator (PI)}
  \footnotetext[4]{Code: \url{github.com/mohammad2012191/VideoAtlas}}
}

\section{Introduction}
\label{sec:intro}

Understanding long-form video requires locating sparse, task-relevant evidence within a massive temporal space: an hour video has 90,000 frames at 25 fps, yet the answer to a query often resides in a few seconds. When a movies editor faces the same challenge, the solution is well-established: a \emph{contact sheet} (a single composite image showing sampled shots) to identify promising regions at a glance before zooming into only those clips. This loop of \emph{overview, identify, zoom} is the key to efficient visual navigation, and it is precisely what current VLMs lack.

\begin{figure}[tb]
  \centering
  \includegraphics[width=1.0\linewidth]{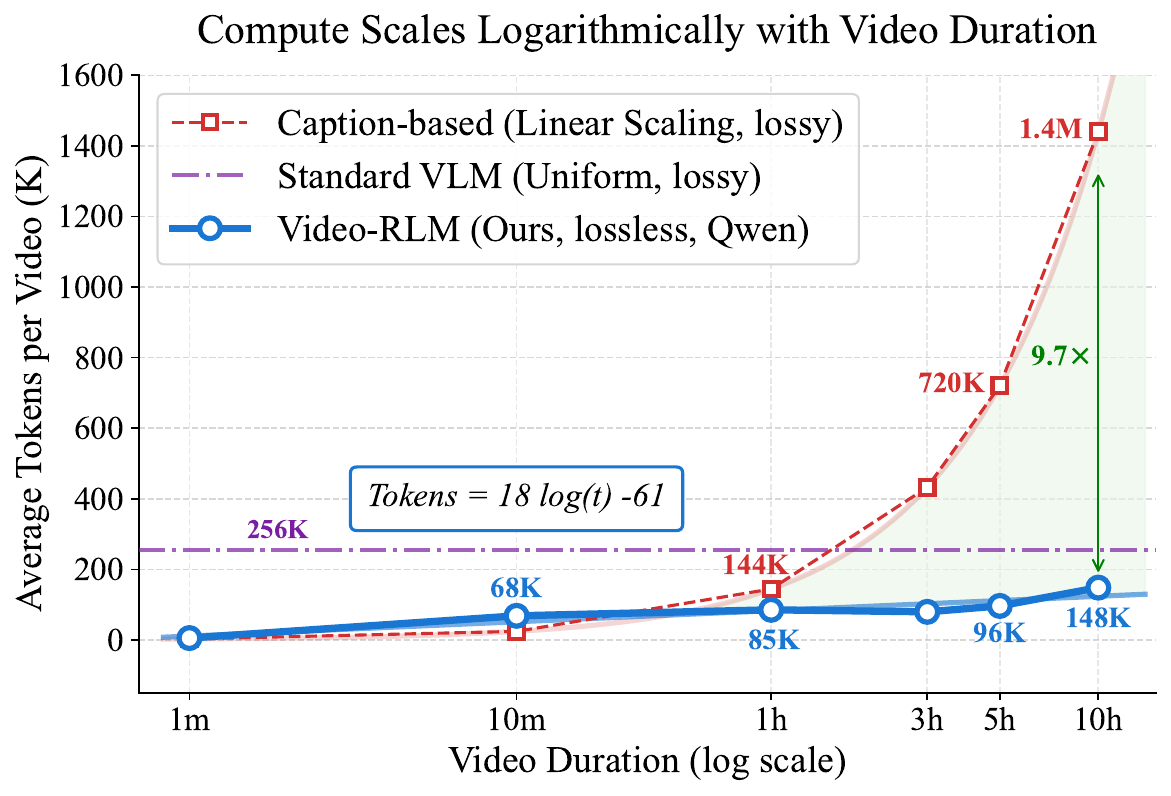}
  \caption{Logarithmic compute scaling with video duration. Video-RLM's hierarchical grid grows sub-linearly ($O(\log T)$), requiring up to 9.7$\times$ fewer tokens than linear-scaling baselines. A uniform VLM maxes out its 256K context trading off sampled frame count with resolution.}
  \label{fig:log_scaling}
\end{figure}

Existing approaches to long-form video understanding can be broadly categorized into four paradigms: uniform sampling, composite grids, caption-based, and agentic-based approaches. Uniform sampling~\cite{wu2024longvideobench,fu2025video} introduces severe temporal sparsity, i.e., at practical budgets, frames are sampled minutes apart, resulting in short events being systematically missed. Moreover, within a fixed context window, increasing the number of sampled frames forces a proportional decrease in per-frame resolution, creating a fundamental coverage-vs-fidelity tradeoff. Composite grids~\cite{kim2024image,eltahir2025vote} pack frames into a single representative image, improving token efficiency but remaining a fixed, lossy snapshot. Caption-based and agentic approaches~\cite{wang2024videoagent,zhang2025deep,yin2025videoarm} rely on text as their primary reasoning medium (captioning clips, storing text summaries, or converting visual observations into language before planning). Even when these systems adaptively sample frames, their intermediate memory and decision-making operate over text, not over a structured visual space. Any visual detail overlooked during transcription or abstraction cannot be recovered by subsequent reasoning. These paradigms also face distinct scalability bottlenecks i.e., standard VLM pipelines~\cite{bai2025qwen3} must decode the video, extract frames, and perform visual tokenization on CPU before any reasoning begins. For long videos, this preprocessing alone can exhaust hundreds of gigabytes of system RAM. Caption-based and agentic methods avoid this by converting video to text first, but incur a different cost: an offline captioning stage that scales linearly with video duration and irreversibly discards visual fidelity. While some agentic methods~\cite{yin2025videoarm} perform this conversion online, they still rely on text as the intermediate representation, inheriting the same information loss.

We claim that a useful video representation must be simultaneously \emph{lossless} (frame-level access at any resolution), \emph{navigable} (agent-directed), \emph{scalable} (no context ceiling), \emph{caption-free} (native visual reasoning), and \emph{preprocessing-free} (no offline decoding). As detailed in \cref{tab:comparison}, current approaches typically optimize for a subset of these properties at the expense of others.

% To the best of our knowledge, no existing representation satisfies even three of these.

\begin{figure*}[t]
  \centering
  \includegraphics[width=\textwidth]{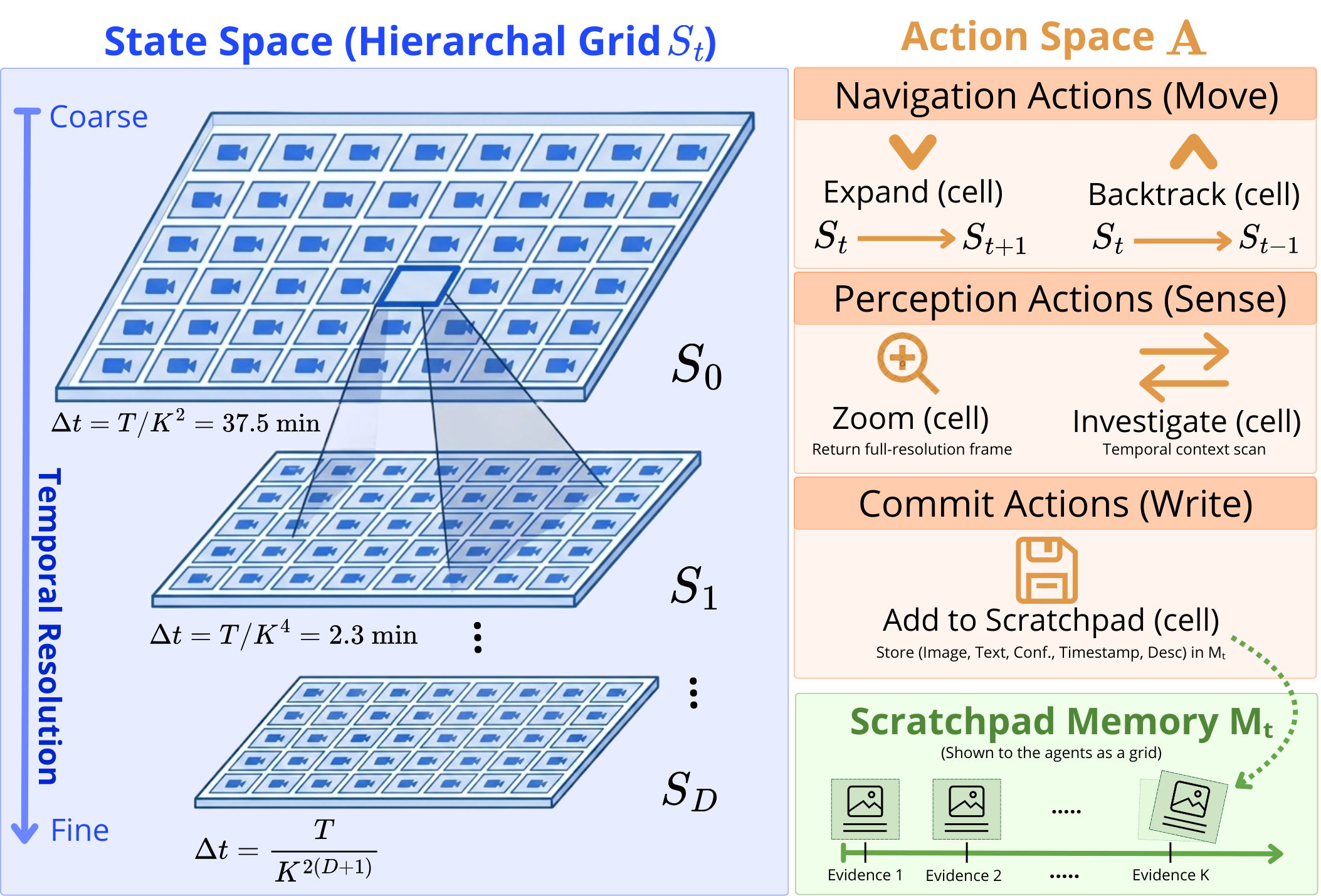}
  \caption{The \textbf{VideoAtlas Environment}. (Left) The state space is a hierarchical grid stack $S_0, S_1, \ldots, S_D$, where $S_0$ is the root grid covering the entire video of duration $T$. Each grid has $K^2$ cells. Deeper levels $d$ provide finer temporal resolution $\Delta t_d = T/K^{2(d+1)}$. (Top Right) The discrete action space $\mathcal{A}$ is divided into navigation (e.g., Expand to $S_{t+1}$), perception, and commit actions. (Bottom Right) The visual scratchpad memory $\mathcal{M}^+$ accumulates multimodal evidence (images, timestamps, QA pairs) across exploration rounds.}

  \label{fig:environment}
\end{figure*}

\emph{VideoAtlas.} We propose a task-agnostic environment that represents any video as a navigable, hierarchical $K \times K$ image grid (\cref{fig:environment}).
The root grid renders the full video as a contact sheet. By invoking \textsc{Expand} (a recursive descent action that generates a new, finer-resolution sub-grid for a selected cell) an agent achieves sub-second temporal precision in $O(\log T)$ steps, where $T$ is the video duration in seconds. The design is uniform throughout: the video, intermediate investigations, and the agent's internal \emph{evidence scratchpad} (a lossless multimodal memory that stores collected frames, subtitles, timestamps, and descriptions) are all rendered as grids. This completely eliminates captioning, offline preprocessing, and context-window ceilings, satisfying all properties in the aforementioned paragraph. Crucially, VideoAtlas also escapes the \emph{coverage-vs-fidelity tradeoff} inherent to uniform VLMs: within a fixed context window, sampling more frames forces lower per-frame resolution, and vice versa. VideoAtlas sidesteps this entirely (each grid image is always rendered at full resolution, and the agent zooms only where needed, never sacrificing visual fidelity for temporal coverage). Structurally, the hierarchy yields \emph{logarithmic} compute growth: as video length increases, only a few additional depth layers are needed rather than linearly more frames. Moreover, the fixed hierarchical grid is inherently \emph{cache-friendly}: root grids and overlapping sub-grids are naturally reused across exploration rounds, achieving 30-60\% multimodal cache hit rates that further reduce effective GPU compute (see Appendix Sec. C.1).

\paragraph{From representation to reasoning.}
With a lossless and navigable video representation in hand, a crucial observation follows: the long-video problem reduces to a \emph{long-context} problem. The video is the context, and what is needed is a mechanism for agents to explore it recursively without compressing it. Recursive Language Models (RLMs)~\cite{zhang2025recursive} provide exactly this mechanism for text, allowing agents to query arbitrarily long contexts through recursive subagent calls and accumulate exact symbolic variables. RLMs, however, require a structured environment to recurse into. VideoAtlas is precisely that structure. We deploy \textbf{Master-Worker Agents} (\textbf{Video-RLM}) within this environment to extend RLMs to the video domain, yielding depth-controlled compute budgeting and logarithmic cost growth. Following are our main contributions.

\begin{enumerate}
    \item \textbf{VideoAtlas.} We formulate video understanding as navigation within a formally defined geometric environment. The hierarchical grid is lossless, caption-free, preprocessing-free,  and strategy-agnostic, with logarithmic access depth, parallelizable subgrids, and a structural cache-friendliness. 

    \item \textbf{Video-RLM.} A parallel Master-Worker architecture extending Recursive Language Models to video. Workers explore grid subtrees concurrently and accumulate evidence in a lossless Visual Scratchpad, while a Master steers exploration via uncertainty analysis.

    \item \textbf{Configurable Traversal Strategies.} Breadth-First and Depth-First instantiations plus a query-adaptive policy that selects traversal order automatically, all composable with the environment without modification.

    \item \textbf{Environment Budgeting.} We budget the \emph{environment}, not the agent: bounding exploration depth $d$ directly controls temporal resolution and compute, providing a principled compute-accuracy hyperparameter.

\end{enumerate}

\noindent Beyond these architectural contributions, experiments reveal that the formulation produces emergent scaling behaviors (adaptive compute allocation and logarithmic cost growth) that we detail in \cref{sec:experiments}.

\makeatletter
\def\thebibliography#1{\section*{\refname\@mkboth{\refname}{\refname}}\small
  \def\list@biblabel##1{##1.}%
  \list{\@biblabel{\arabic{enumiv}}}%
       {\settowidth\labelwidth{\@biblabel{#1}}%
        \leftmargin\labelwidth
        \advance\leftmargin\labelsep
        \@openbib@code
        \usecounter{enumiv}%
        \let\p@enumiv\@empty
        \renewcommand\theenumiv{\arabic{enumiv}}}%
  \sloppy\clubpenalty4000\widowpenalty4000%
  \sfcode`\.\@m} % \clearpage was removed from this line
\makeatother

% ================================================================
% 2. RELATED WORK
% ================================================================
\section{Related Work}
\label{sec:related}

\paragraph{Long-Form Video Understanding.}
Standard Video-Language Models process videos by uniformly sampling a fixed number of frames in a single forward pass~\cite{wu2024longvideobench,fu2025video}. This introduces two structural problems. First, at any practical budget (e.g., 64 frames in an hour), the temporal stride is $\sim$56 seconds per frame, so short events, fine-grained visual details, and scene transitions are easily missed. Second, the context window imposes a hard ceiling: beyond a few hundred high-resolution frames, the model truncates input or degrades. One practical workaround is to pack multiple frames into a single $K\times K$ composite image (a contact-sheet grid)~\cite{kim2024image,eltahir2025vote}, which improves token efficiency. However, a single-resolution grid is still fundamentally \emph{lossy}: it represents the video with a fixed sample of moments and cannot recover the events in between. Grids alleviate the context-packing problem, but they do not resolve the \emph{coverage} problem.

\paragraph{Caption-Based Approaches.}
A prominent line of work avoids the frame-count limit by first transcribing the video into text captions and then reasoning over them.
LLoVi~\cite{zhang2024simple} converts densely sampled short clips into text summaries and aggregates them with an LLM. MR.Video~\cite{pangmr} scales this with a MapReduce design: clips are captioned in parallel, standardized, and then synthesized into a final answer by a reducer LLM. Video to Text conversion is standard practice, although systems that explicitly observe video frames at a coarse step immediately convert those observations into text before any planning or memory update. Pang~\etal~\cite{pangmr} explicitly acknowledge that video-to-text modality transitions cause reasoning failures on scene transitions and fine-grained visual details.
\paragraph{Agentic, Hierarchical, and Memory Approaches.}
Another set of approaches treat long-video understanding as agentic search. DVD~\cite{zhang2025deep} constructs a multi-granular database (global summaries, clip captions/embeddings, and indexed raw frames) and queries it with tools (Global Browse, Clip Search, Frame Inspect). VideoARM \cite{yin2025videoarm} performs on-the-fly coarse-to-fine search via a set of predefined tools (e.g., captioning, temporal localization, visual QA) over a hierarchical multimodal memory, avoiding exhaustive preprocessing. VideoTree~\cite{wang2025videotree} builds a query-adaptive hierarchical representation to guide efficient exploration. On the memory side, WorldMM~\cite{yeo2025worldmmdynamicmultimodalmemory} organizes long-video memory into episodic, semantic, and visual components, retrieved adaptively per query~\cite{hu2024visual}.

Despite their diversity, these systems share a common limitation: intermediate evidence is stored as captions, text summaries, or compressed embeddings, never as raw visual frames, meaning none provide lossless, navigable access to any arbitrary video moment by construction.

\begin{table}[tb]
  \caption{Comparison of long-video QA methods. ``Caption-Free'' = no text captions used as intermediate representation. ``Lossless'' = no information lost between input and reasoning. ``$\infty$ Context'' = can handle arbitrarily long videos without context overflow. ``Parallel'' = workers explore concurrently.}
  \label{tab:comparison}
  \centering
  \small
  \setlength{\tabcolsep}{4pt}
  \begin{tabular}{@{}lcccc@{}}
    \toprule
    Method & Caption-Free & Lossless & $\infty$ Context & Parallel \\
    \midrule
    VLM (Uniform) & \cmark & \xmark & \xmark & \xmark \\
    LLoVi~\cite{zhang2024simple} & \xmark & \xmark & \xmark & \xmark \\
    MR.Video~\cite{pangmr} & \xmark & \xmark & \xmark & \cmark \\
    VideoARM~\cite{yin2025videoarm} & \xmark & \xmark & \xmark & \xmark \\
    DVD~\cite{zhang2025deep} & \xmark & \xmark & \xmark & \xmark \\
    \midrule
    \textbf{Ours} & \cmark & \cmark & \cmark & \cmark \\
    \bottomrule
  \end{tabular}
\end{table}

\paragraph{Long Context as the Core Challenge.}
Recursive Language Models (RLMs)~\cite{zhang2025recursive} address long text contexts by letting agents access context through recursive subagent calls, storing results in lossless symbolic variables rather than compressing them into the model's context window. The RLM insight transfers naturally to video, but only if an \emph{environment} is defined in which agents can navigate the video visually. Existing video ``environments'' are built around clip databases and text-based retrieval~\cite{zhang2025deep,yin2025videoarm}. No visual, lossless, recursively navigable environment for video has been proposed. VideoAtlas fills precisely this gap.

% --- Figure 2: The VideoAtlas Environment ---
% --------------------------------------------

\paragraph{Environment Budgeting vs. Prior Compute Adaptation.}
Chain-of-thought reasoning~\cite{wei2022chain} and adaptive test-time compute allocation~\cite{snell2025scaling} have shown that allocating more inference compute consistently improves performance on language and reasoning tasks. In the video domain, the closest analog is VideoARM~\cite{yin2025videoarm}, which adaptively chooses how many frames $N_1$ to sample per localized interval, a form of density adaptation that improves efficiency. However, this controls sampling \emph{quantity} (how many frames), not structural \emph{resolution} (how fine the temporal decomposition is): within each interval, sampling remains uniform, and events falling between sample points can still be missed regardless of $N_1$. MR.Video~\cite{pangmr} offers no such control at all. Its captioning cost is fixed by video duration regardless of the query. A fundamentally different form of budgeting is absent from prior works: controlling the \emph{temporal resolution of the environment itself}, where each depth level geometrically subdivides time, providing formal precision guarantees calibrated to video length and query granularity. We introduce exactly this form of budgeting with VideoAtlas.

\paragraph{What Is Missing?}
\cref{tab:comparison} summarizes the key properties of representative methods. In the next section, we introduce VideoAtlas, which addresses all the aforementioned gaps.

% ================================================================
% 3. METHODOLOGY
% ================================================================
\section{Methodology}
\label{sec:environment}

\begin{figure*}[t]
  \centering
  \includegraphics[width=0.85\textwidth]{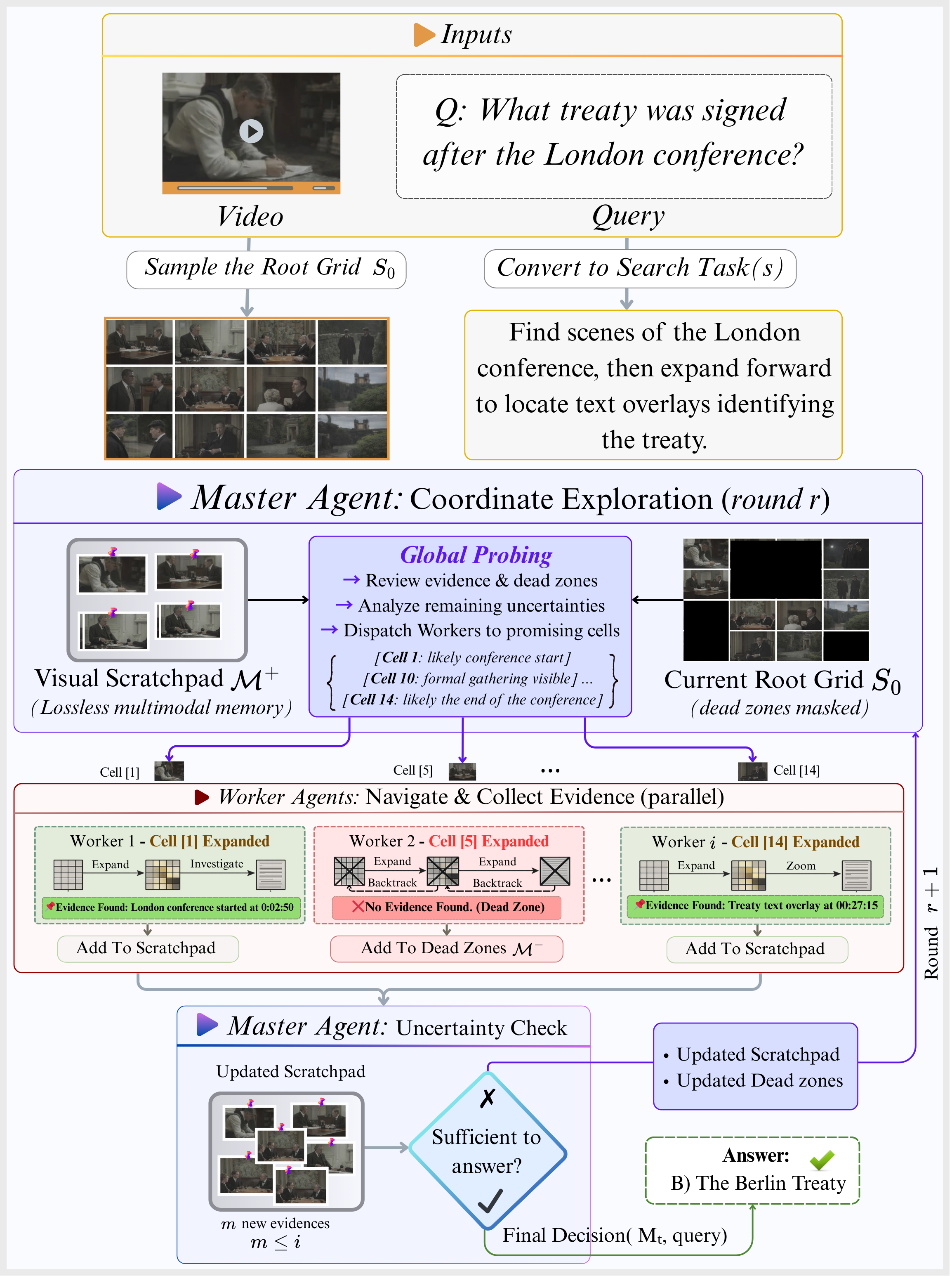}
  \caption{\textbf{Video-RLM} overview. The query is converted into a search task. In each round $r$, the Master examines the root grid $S_0$ (with dead zones masked) and the scratchpad $\mathcal{M}^+$, then assigns promising cells to Workers. Each Worker autonomously explores its assigned region via navigation, perception, and commit actions. After all Workers return, $\mathcal{M}^+$ and $\mathcal{M}^-$ are updated. The Master performs an uncertainty analysis: if evidence is sufficient, the final answer is produced. Otherwise, a new round begins.}
  \label{fig:overview}
\end{figure*}

We present our methodology in two parts. First, we introduce \textbf{VideoAtlas} (\cref{sec:videoatlas}): a task-agnostic environment that renders any video as a navigable, hierarchical grid with formally defined state, action, and observation spaces (\cref{fig:environment}). Second, we describe \textbf{Video-RLM} (\cref{sec:rlm}): a parallel Master-Worker agent architecture that operates within VideoAtlas to answer questions about arbitrarily long videos (\cref{fig:overview}).

% ================================================================
% --- Figure 3: Video-RLM Overview ---
% ------------------------------------

\subsection{VideoAtlas}
\label{sec:videoatlas}

\paragraph{Hierarchical Grid.}
At the core of VideoAtlas is a recursive $K \times K$ image grid (default $K{=}8$, yielding 64 cells). Given a video of duration $T$ seconds, the root grid $S_0$ assigns each cell $c_i$ to a contiguous temporal interval $[t_i^{\text{start}}, t_i^{\text{end}}]$ and displays a representative frame sampled at the interval midpoint, providing a ``bird's-eye view'' of the entire video (\cref{fig:overview}). Every cell is \emph{addressable}: applying \textsc{Expand} to cell $c_i$ deterministically generates a child grid $S_{d+1}$ for that cell's sub-interval, increasing temporal resolution by a factor of $K^2$. At depth $d$, the temporal resolution is $\Delta t_d = T / K^{2(d+1)}$, and reaching any frame requires at most $D_{\max} = \lceil\log_{K^2}(T\cdot\mathrm{fps})\rceil$ steps, achieving sub-second precision even for 10-hour videos. Sub-grids are generated on-the-fly with no offline preprocessing. Agents interact with raw frames at every level.

\paragraph{Action Space.}
Unlike agentic methods~\cite{yin2025videoarm} whose actions perform \emph{video-processing} operations (captioning, translating), VideoAtlas exposes \emph{environment-navigation} actions grouped into three categories (\cref{fig:environment}, right):

\textbf{Navigation} (move through the hierarchy): \textsc{Expand}$(c_i)$ descends into cell $c_i$, generating a child grid. \textsc{Backtrack}$()$ returns to the parent grid. \textsc{MarkPromising}$(c_i, c_j, \ldots)$ flags cells for later exploration via a FIFO queue (BFS mode only).

\textbf{Perception} (sense the environment): \textsc{Zoom}$(c_i)$ returns a full-resolution frame for cell $c_i$. \textsc{Investigate}$(c_i, \text{direction (before/after)})$ generates a temporal context scan of the frames immediately before or after a cell, used when an anchor event is found but the answer lies in neighboring frames.

\textbf{Commit} (record evidence): \textsc{AddToScratchpad}$(\text{items})$ stores evidence tuples to the scratchpad. \textsc{Finished}$()$ declares the current region fully explored.

The available action set is state-dependent: \textsc{Expand} is removed when cell span drops below a threshold (e.g., ${<}1$\,s), \textsc{Backtrack} is removed at the root, and BFS and DFS workers receive different action sets. \textbf{The agent cannot select what it cannot see}, eliminating invalid actions by construction, while deciding its own explore-exploit balance from visual cues.

\paragraph{Memory.}
\textbf{Positive memory} ($\mathcal{M}^+$, Visual Scratchpad): a lossless multimodal memory that stores evidence as tuples $(I_{\text{img}}, s, \tau, c, d)$ representing image patch, subtitle, timestamp, confidence score, and a text description relating the evidence to the query. When presented to the VLM, $\mathcal{M}^+$ is rendered as a grid image with timestamps, subtitles, and indices burned into pixel space, enabling unambiguous cross-referencing.
\textbf{Negative memory} ($\mathcal{M}^-$, Dead Zones): intervals explored with no relevant findings are marked as dead zones. The grid renderer enforces this \emph{visually} by blacking out overlapping cells, physically preventing the VLM from hallucinating details in already-explored regions.

\paragraph{Formal Environment Definition.}
\label{sec:properties}
At any step, the environment state $S$ comprises five components: the current temporal position $p = (\text{center}, \text{span})$, the depth $d$ in the hierarchy, the positive and negative memories $\mathcal{M}^+$ and $\mathcal{M}^-$, and the navigation stack $\sigma$ for backtracking. The observation is the grid image rendered for the interval defined by $p$ at depth $d$, together with aligned subtitle context filtered for the current temporal window.

This state definition, together with the action space, formally defines a Markov Decision Process (MDP). The reward is task-defined (e.g., answer correctness for QA, temporal IoU for grounding) making VideoAtlas a general substrate for any task reducible to ``find relevant moments in a video.'' In this work we solve it via zero-shot VLM reasoning, but the formal MDP opens a direct path to reinforcement learning. The environment exhibits four structural properties: (1)~\emph{Parallelizable}: the grid decomposes into independent subtrees explorable concurrently. (2)~\emph{Traversal-agnostic}: BFS, DFS, beam search, or learned policies can govern expansion order without modifying the environment. (3)~\emph{Depth-controlled compute}: bounding $d$ yields a principled compute-accuracy hyperparameter. (4)~\emph{Logarithmic overhead}: as video duration grows, the hierarchy adds depth levels logarithmically, yielding $\mathcal{O}(\log T)$ scaling rather than $\mathcal{O}(T)$. Notably, the depth parameter $d$ interpolates between uniform sampling ($d{=}0$, equivalent to a single composite grid of $K^2$ frames) and full recursive exploration ($d{=}D_{\max}$); prior uniform-sampling and composite-grid methods are thus degenerate cases of VideoAtlas with no exploration.

% ================================================================
\subsection{Video-RLM: Master-Worker Architecture}
\label{sec:rlm}

We extend Recursive Language Models~\cite{zhang2025recursive} to videos by deploying agents in VideoAtlas. Agents access video context through recursive subagents (workers) and store outputs in the Visual Scratchpad $\mathcal{M}^+$. Exploration proceeds in discrete \emph{rounds}: in each round $r$, the Master assigns cells to Workers, Workers explore in parallel, and results are merged into $\mathcal{M}^+$ and $\mathcal{M}^-$ before the next round begins (\cref{fig:overview}).

\paragraph{Search Task Extraction.}
Before visual exploration, a text-only step converts the raw query into a concrete search task. For example: \emph{``What treaty was signed after the London conference?''} $\rightarrow$ \emph{``Find the London conference scene. Look immediately after for treaty names in text overlays or subtitles.''} This search task guides all subsequent prompts.

\paragraph{Master Agent.}
The Master holds the global view: it examines the root grid $S_0$ (with dead zones masked) and the current scratchpad $\mathcal{M}^+$, then selects promising cells for the next round (\emph{Global Probing}). A priority queue with \emph{Virtual Loss}~\cite{chaslot2008parallel} ensures that cells already assigned to workers are deprioritized, preventing redundant exploration. After each round, the Master performs \emph{Uncertainty Analysis}: (a)~a sufficiency check, (b)~temporal interpolation to suggest targeted search bounds from gaps between evidence anchors, and (c)~dynamic memory pruning.

\paragraph{Worker Agents.}
Each worker receives one cell from the frontier and explores it autonomously. Two modes are supported: \textbf{Depth-First Search (DFS) mode} where the worker \textsc{Expand}s deeper into the timeline with a multi-step budget, ideal for localizing specific details. \textbf{Breadth-First Search (BFS) mode} where the worker scans one level with a single-step budget, ideal for evidence spread across the video. The traversal queue is re-prioritized via the Master's visual scoring.

\paragraph{Query-Adaptive Traversal.}
The Master selects the traversal strategy before any frames are processed by analyzing the query's linguistic traits: DFS for specific detail localization, BFS for sequence or flow understanding.

\paragraph{Sufficiency, Stopping, and Final Decision.}
\label{sec:final}
Exploration stops at three levels: (1)~worker-level (budget exhausted or \textsc{Finished}), (2)~master-level (sufficiency check passes after round $r$), (3)~global (total compute budget reached). Once exploration terminates, the Master synthesizes the answer from $\mathcal{M}^+$: it sees the actual collected evidence frames (rendered as a grid with burned-in labels), not text summaries, and evaluates each candidate against the visual evidence.

% ================================================================
% 6. EXPERIMENTS
% ================================================================
\section{Experiments}
\label{sec:experiments}

% ----------------------------------------------------------------
\subsection{Experimental Setup}

\paragraph{Benchmarks.}
We evaluated VideoRLM on the \textbf{long} subsets of two benchmarks: LongVideoBench~\cite{wu2024longvideobench} (LVB, 15-60\,min videos) and Video-MME~\cite{fu2025video} (VMME, without subtitles). To stress-test scalability beyond VLM context limits, we constructed \textbf{10-hour variants} by concatenating multiple videos from each benchmark. Each query targeted a single source video placed at a random position among distractors. Subtitle tracks were merged with correct temporal offsets. This isolates the ``needle in a haystack'' challenge at extreme durations. For VMME, we evaluated the system \emph{without subtitles} to verify that the system genuinely understands visual content rather than relying on textual cues.

\paragraph{\textbf{Model}.}
Our primary experiments used Qwen3.5-35B-A3B~\cite{qwen3.5} (35B total parameters, 3B active per forward pass) for both Master and Workers, differentiated via separate system prompts and action sets, served through vLLM~\cite{kwon2023efficientmemorymanagementlarge} on 4$\times$A100 80\,GB GPUs. Each image in the grids used throughout the system was rendered at a unified resolution of $320 \times 320$ pixels. We use grid size $K=8$. To demonstrate VLM-agnosticism, we additionally evaluate with Gemini-3-Flash as the backbone (\cref{tab:qa_main}).

\paragraph{\textbf{Baselines}.}
We compared against five categories. (1)~\textbf{Proprietary Models}: GPT-4o, GPT-5, Gemini-3-Flash, and Claude-Opus-4.5. (2)~\textbf{Open-Source VLMs} InternVL3.5 241B (28B active) and GLM-4.5V-106B (12B active), uniform-sampling baselines with significantly larger active parameters than ours. (3)~\textbf{Caption Reliant Agentic Methods}: DVD~\cite{zhang2025deep}, MR.Video~\cite{pangmr}, and VideoARM~\cite{yin2025videoarm}, reported from their original papers. (4)~\textbf{Uniform Sampling}: Qwen3.5-35B-A3B with 160 uniformly sampled frames at a resolution of $320 \times 320$ pixels (similar to our framework) along with their temporally aligned subtitles, representing the strongest single-pass baseline within our hardware budget. (5)~\textbf{LLM over Captions}: following LLoVi~\cite{zhang2024simple}, GPT-4o captions (from the MR.Video repository) are concatenated temporally and answered by Qwen3.5-35B-A3B, isolating the benefit of visual exploration over textual summarization.

% ----------------------------------------------------------------
\subsection{Main Results}

\cref{tab:qa_main} compares all methods on the standard long-video benchmarks.
\cref{tab:qa_10hr} evaluates the 10-hour extended-duration variants, alongside average token consumption per question.

% ----- Table 1: Standard Benchmarks -----
\begin{table}[tb]
  \caption{%
    Video QA accuracy (\%) on the standard (Long) subsets.
    LVB: LongVideoBench. VMME: Video-MME (no subs).
  }
  \label{tab:qa_main}
  \centering
  \small
  \setlength{\tabcolsep}{4pt}
  \renewcommand{\arraystretch}{1.08}
  \begin{tabular}{@{}lc cc@{}}
    \toprule
    Method & Active & LVB & VMME \\
    \midrule
    \rowcolor{gray!8}
    \multicolumn{4}{l}{\textit{Proprietary Models}} \\
    GPT-4o~\cite{hurst2024gpt}          & Prop. & 66.7 & 65.3 \\
    GPT-5~\cite{singh2025openai}                     & Prop. & 72.6 & \textbf{81.8} \\
    Gemini-3-Flash~\cite{team2023gemini}& Prop. & 74.5 & ---  \\
    Claude-Opus-4.5~\cite{anthropic2024claude} & Prop. & 67.2 & 77.6 \\
    \midrule
    \rowcolor{gray!8}
    \multicolumn{4}{l}{\textit{Open-Source VLMs}} \\
    InternVL3.5-241B~\cite{chen2024internvl}  & 28B & 67.1 & 72.9 \\
    GLM-4.5V-106B~\cite{glm2024chatglm}      & 12B & \textbf{76.7} & 74.6 \\
    \midrule
    \rowcolor{gray!8}
    \multicolumn{4}{l}{\textit{Caption-Reliant Agentic Methods}} \\
    MR.Video~\cite{pangmr} (Gemini+GPT-4o)             & Prop. & 61.6 & 61.8 \\
    VideoARM~\cite{yin2025videoarm} (Qwen3-VL-235B)     & 22B   & ---  & 54.9 \\
    VideoARM~\cite{yin2025videoarm} (GPT-o3+GPT-4o)     & Prop. & 76.4 & 81.2\\
    \midrule
    \rowcolor{gray!8}
    \multicolumn{4}{l}{\textit{Same-Backbone Baselines}} \\
    Qwen3.5 (uniform, 160 fr.)            & 3B              & 61.5 & 63.8 \\
    LLM over Captions$^{\star}$         & Prop.{\tiny+}3B & 62.4 & 64.2 \\
    \midrule
    \rowcolor{gray!8}
    \multicolumn{4}{l}{\textit{VideoAtlas (Ours)}} \\
    \rowcolor{yellow!15}
    \textbf{Video-RLM (Qwen3.5-35B)}    & \textbf{3B}    & 52.5 & 50.4 \\
    \rowcolor{yellow!15}
    \textbf{Video-RLM (Gemini-3-Flash)} & \textbf{Prop.} & 72.0 & 76.2 \\
    \bottomrule
  \end{tabular}
  \smallskip
  \par\noindent{\small $^{\star}$\,GPT-4o captions (MR.Video repo), answered by Qwen3.5-35B-A3B.}
\end{table}

\begin{table}[tb]
  \caption{%
    10-hour variant: accuracy (\%) and average tokens per question. {\color{red}$\Delta$}: accuracy drop from standard benchmarks (\cref{tab:qa_main}).
  }
  \label{tab:qa_10hr}
  \centering
  \small
  \setlength{\tabcolsep}{8pt}
  \renewcommand{\arraystretch}{1.08}
  \begin{tabular}{@{}l ccc@{}}
    \toprule
    Method & Acc. & {\color{red}$\Delta$} & Avg. Tokens \\
    \midrule
    \rowcolor{gray!8}
    \multicolumn{4}{l}{\textbf{LVB-10hr}} \\
    Qwen3.5 (uniform)       & 49.2 & -12.3 & 212K \\
    LLM over Captions$^{*}$ & 62.1 & \textbf{-0.3}  & 207K$^{\ddagger}$ \\
    \rowcolor{yellow!15}
    \textbf{Video-RLM (Qwen)}   & 47.7 & -4.8  & 146K$^{\dagger}$ \\
    \rowcolor{yellow!15}
    \textbf{Video-RLM (Gemini)} & \textbf{70.1} & -1.9  & 307K \\
    \midrule
    \rowcolor{gray!8}
    \multicolumn{4}{l}{\textbf{VMME-10hr (no subs)}} \\
    Qwen3.5 (uniform)       & 50.6 & -13.2 & 232K \\
    LLM over Captions$^{*}$ & 36.0 & -28.2 & 235K$^{\ddagger}$ \\
    \rowcolor{yellow!15}
    \textbf{Video-RLM (Qwen)}   & 49.7 & \textbf{-0.7}  & 403K$^{\dagger}$ \\
    \rowcolor{yellow!15}
    \textbf{Video-RLM (Gemini)} & \textbf{69.1} & -7.1  & 390K \\
    \bottomrule
  \end{tabular}
  \smallskip
  \par\noindent{\small
  $^{*}$\,GPT-4o captions + Qwen3.5 (exceeds 256K context in many samples).
  $^{\dagger}$\,Effective tokens after vLLM multimodal prefix cache (avg.\,36-42\% hit rate).
  $^{\ddagger}$\,QA tokens only, excludes GPT-4o captioning cost.}
\end{table}

\paragraph{Standard benchmarks.}
At standard durations (\cref{tab:qa_main}), Video-RLM (3B active parameters) achieves accuracy competitive with substantially larger open-source VLMs (12-28B active) and proprietary baselines. Importantly, Qwen 3.5 is video-finetuned, whereas Video-RLM assumes a purely \emph{zero-shot} agent with no video-specific training, achieving these results without any intermediate text representation or captioning. The accuracy gap between zero-shot navigation and finetuned uniform sampling narrows with a stronger backbone: Video-RLM (Gemini) reaches 72.0\% on LVB, within 2.5 points of Gemini-3-Flash's direct performance (74.5\%), confirming VideoAtlas is VLM-agnostic and performance scales with backbone capability.

\paragraph{Extended duration (10 hours).}
The 10-hour variants (\cref{tab:qa_10hr}) reveal a more significant comparison. LLM over Captions \emph{collapses to 36.0\%} on VMME-10hr: concatenated captions exceed Qwen's 256K context window, forcing truncation and information loss, demonstrating linear captioning fails beyond context limits. Notably, captions degrade only $-0.3\%$ on LVB-10hr (where subtitle tracks are available) but $-28.2\%$ on VMME-10hr (no subtitles), exposing a heavy reliance on textual cues rather than genuine visual understanding.
Uniform Qwen degrades moderately (63.8\%$\to$50.6\% VMME) as sampling becomes prohibitively sparse. Video-RLM maintains highly stable performance across durations (e.g., Qwen VMME drops only 0.7\% vs uniform 13.2\% and Captions 28.2\%), validating that VideoAtlas buffers the agent against duration scaling. On VMME-10hr, the absence of subtitles forces the purely zero-shot Qwen agent into extensive visual exploration (403K effective tokens), while the stronger Gemini backbone zeroes in on the answer faster (390K tokens), an emergent adaptive compute property where weaker perception necessitates more search steps. Crucially, the recursive nature of the environment is inherently \emph{cache-friendly}: workers re-examine the same grid view across multiple reasoning steps that do not change the navigation state, creating repeated visual token prefixes. For self-hosted Qwen (vLLM), automatic multimodal prefix caching exploits this redundancy transparently, achieving 36-42\% hit rates at 10 hours (up to 61\% for shorter videos, see Appendix Sec. C1). Video-RLM (Gemini) achieves 70.1\% on LVB-10hr with near-zero degradation (-1.9\%) from the standard benchmark.

\paragraph{Error analysis.}
Manual inspection of failure cases reveals three dominant error modes: VLM perception errors (misreading text overlays, confusing visually similar scenes), premature sufficiency (the Master declares evidence sufficient despite contradictions), and text latching (anchoring on phrases in evidence that superficially match a candidate answer). All three are model-dependent, as confirmed by the substantial accuracy improvement when switching from Qwen (3B active) to Gemini-3-Flash without any changes to VideoAtlas. We discuss these errors in more detail in the Appendix Sec. A.

\subsection{Logarithmic Compute Scaling}
\label{sec:log_scaling}
\cref{fig:log_scaling} demonstrates the fundamental efficiency advantage of hierarchical navigation using LVB-10hr. As video duration grows from 1 minute to 10 hours (a 600$\times$ increase), Video-RLM's compute cost increases sub-linearly: the hierarchy adds depth levels logarithmically ($\lceil\log_{K^2}(T\cdot\text{fps})\rceil$), and the sufficiency mechanism halts exploration once evidence is found. Caption-based pipelines scale linearly, where every clip must be captioned regardless of the query, requiring over 1.4M tokens per query at 10 hours. Video-RLM achieves comparable accuracy using only 148K effective tokens (a 9.7$\times$ reduction), and unlike uniform VLMs whose cost is fixed, depth can always be extended to accommodate longer videos.

% ----------------------------------------------------------------
\subsection{Environment Budgeting and Adaptive Compute}

\begin{figure}[tb]
  \centering
  \includegraphics[width=\linewidth]{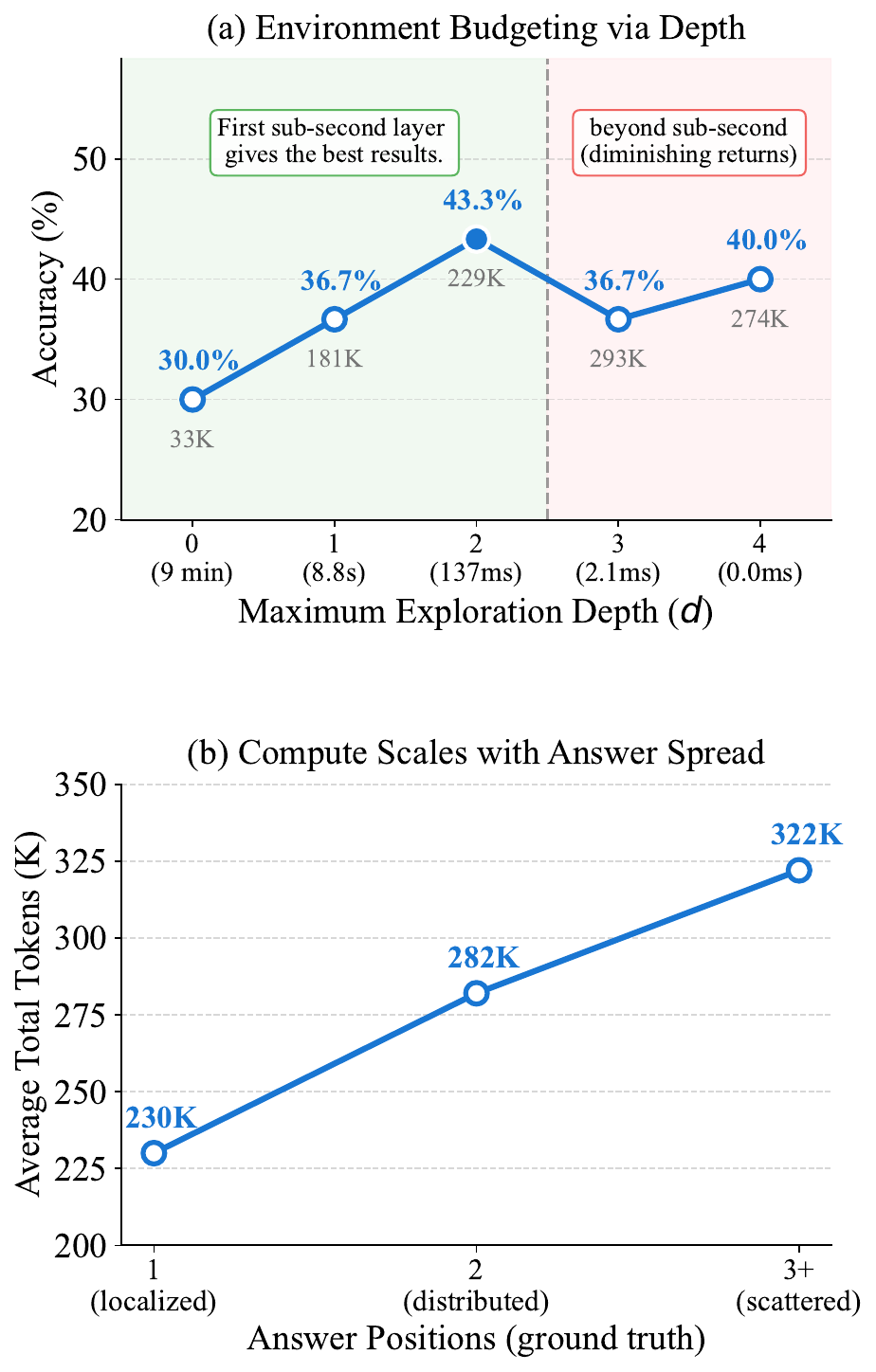}
  \caption{(a)~Environment budgeting: accuracy and tokens vs.\ max depth on subset of LVB-10hr (temporal span annotated). Green: optimal depth (first sub-second layer). (b)~Adaptive compute: average tokens scale with evidence spread without ground-truth supervision.}
  \label{fig:depth_and_adaptive}
\end{figure}

\cref{fig:depth_and_adaptive}(a) shows accuracy vs.\ maximum exploration depth on 30 questions sampled from LVB-10hr. Accuracy rises from 30\% ($d{=}0$, root grid only) to 43.3\% ($d{=}2$, 137\,ms span), then plateaus at depths 3-4 where the finest resolution drops below one millisecond, well beyond any meaningful visual granularity. In practice, we set the maximum depth to the first sub-second layer, which automatically adapts to video duration (a 1-minute video reaches sub-second at $d{=}1$, a 10-hour video at $d{=}2$). Depth $d$ is thus a principled compute-accuracy hyperparameter that directly controls temporal \emph{resolution} rather than frame \emph{quantity}.

\cref{fig:depth_and_adaptive}(b) verifies that compute allocation adapts to question difficulty without explicit supervision. Grouping LVB questions by the number of ground-truth temporal positions containing answer evidence, scattered answers (3+ positions) consume 40\% more tokens (322K vs.\ 230K) than localized ones. This emergent behavior arises from the interaction between the Master's uncertainty analysis, the sufficiency mechanism, and the hierarchical structure.

% ----------------------------------------------------------------
% ----------------------------------------------------------------
\subsection{Worker Scaling}

\begin{figure}[tb]
  \centering
  \includegraphics[width=1.0\linewidth]{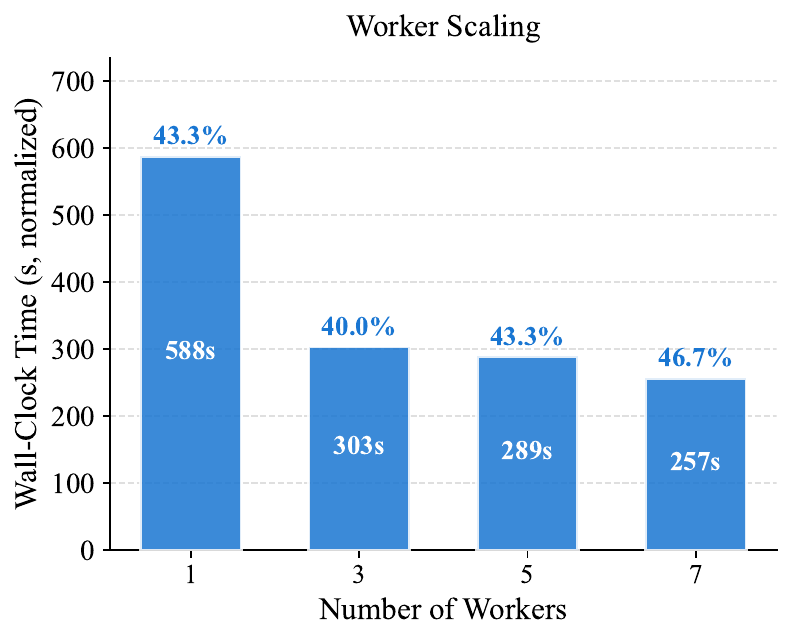}
  \caption{Wall-clock time (normalized to equal workload) vs.\ number of workers 30 questions sampled from LVB-10hr. Accuracy (annotated) remains stable across all configurations.}
  \label{fig:parallelism}
\end{figure}

We vary the number of workers $\in \{1, 3, 5, 7\}$ on 30 questions sampled from LVB-10hr (\cref{fig:parallelism}). After normalizing for workload differences, wall-clock time decreases from 588s (1 workers) to 257s (7 workers), a 2.25$\times$ speedup, while accuracy remains stable (40-47\%). This is a structural property of the environment: each subtree is self-contained, so adding workers improves throughput without modifying the search protocol.

\section{Limitations}
\label{sec:limitations}

We identify four principal limitations.
\emph{(1)~VLM perception bottleneck}: the system's perceptual ceiling is set entirely by the backbone VLM. Our error analysis reveals three dominant failure modes, all VLM-dependent rather than environment-dependent: (a)~perception errors (misreading text overlays, confusing visually similar scenes), (b)~premature sufficiency, where the Master declares evidence sufficient despite contradictions rather than directing further exploration, and (c)~text latching, where the agent over-relies on subtitle cues when available. Crucially, switching from Qwen (3B active) to Gemini-3-Flash eliminates a substantial portion of these errors without any architectural changes (\cref{tab:qa_main}), confirming that performance scales directly with backbone capability and will improve as VLMs advance.
\emph{(2)~No-anchor exploration overhead}: when the root grid $S_0$ contains no visually obvious anchor for the query, the agent may require additional exploration rounds before finding relevant regions. The Master mitigates this progressively, as each round's newly collected evidence refines subsequent cell assignments. Developing methods to surface semantically relevant information into upper depth layers could substantially improve efficiency and is a promising direction for future work.
\emph{(3)~Evaluation scope}: we validate on multiple-choice QA. The MDP formulation supports temporal grounding, summarization, and anomaly detection (only the reward signal changes. The environment does not), but these remain to be demonstrated empirically.
\emph{(4)~Zero-shot only}: we solve the MDP entirely via zero-shot VLM reasoning, the weakest possible agent. The discrete, finite action space makes VideoAtlas directly compatible with RL training (PPO, DQN), which would likely improve exploration efficiency, but we leave this to future work.

% ================================================================
% 8. CONCLUSION
% ================================================================
\section{Conclusion}
\label{sec:conclusion}

We introduced VideoAtlas, a formulation that reframed video understanding as navigation within a formally defined hierarchical environment, and Video-RLM, a parallel Master-Worker agent that operates within it. Three properties emerged from the formulation: logarithmic compute growth with duration, principled environment budgeting via depth control, and emergent adaptive compute allocation. This environment defines the state, action, and observation spaces of a complete MDP opening a direct path from zero-shot reasoning to learned exploration policies, and from question answering to any task reducible to ``find relevant moments in a video.''

\section*{Acknowledgements}
We are grateful to the KAUST Academy for its generous support, and especially to Prof.\ Sultan Albarakati that made this work possible. For computer time, this research used Ibex managed by the Supercomputing Core Laboratory at King Abdullah University of Science \& Technology (KAUST) in Thuwal, Saudi Arabia.

% ================================================================
% REFERENCES
% ================================================================
\bibliographystyle{splncs04}
\bibliography{main}
\clearpage
\input{appendix}

\end{document}

%% file: appendix.tex
% ================================================================
% SUPPLEMENTARY MATERIAL
% ================================================================
\clearpage
\setcounter{section}{0}
\renewcommand{\thesection}{\Alph{section}}

\section*{Appendix}

% ================================================================
% REASONING TRACE EXAMPLE
% ================================================================

% ================================================================
% REASONING TRACE EXAMPLE
% ================================================================

\begin{figure}[h]
  \centering
  \includegraphics[width=0.9\linewidth]{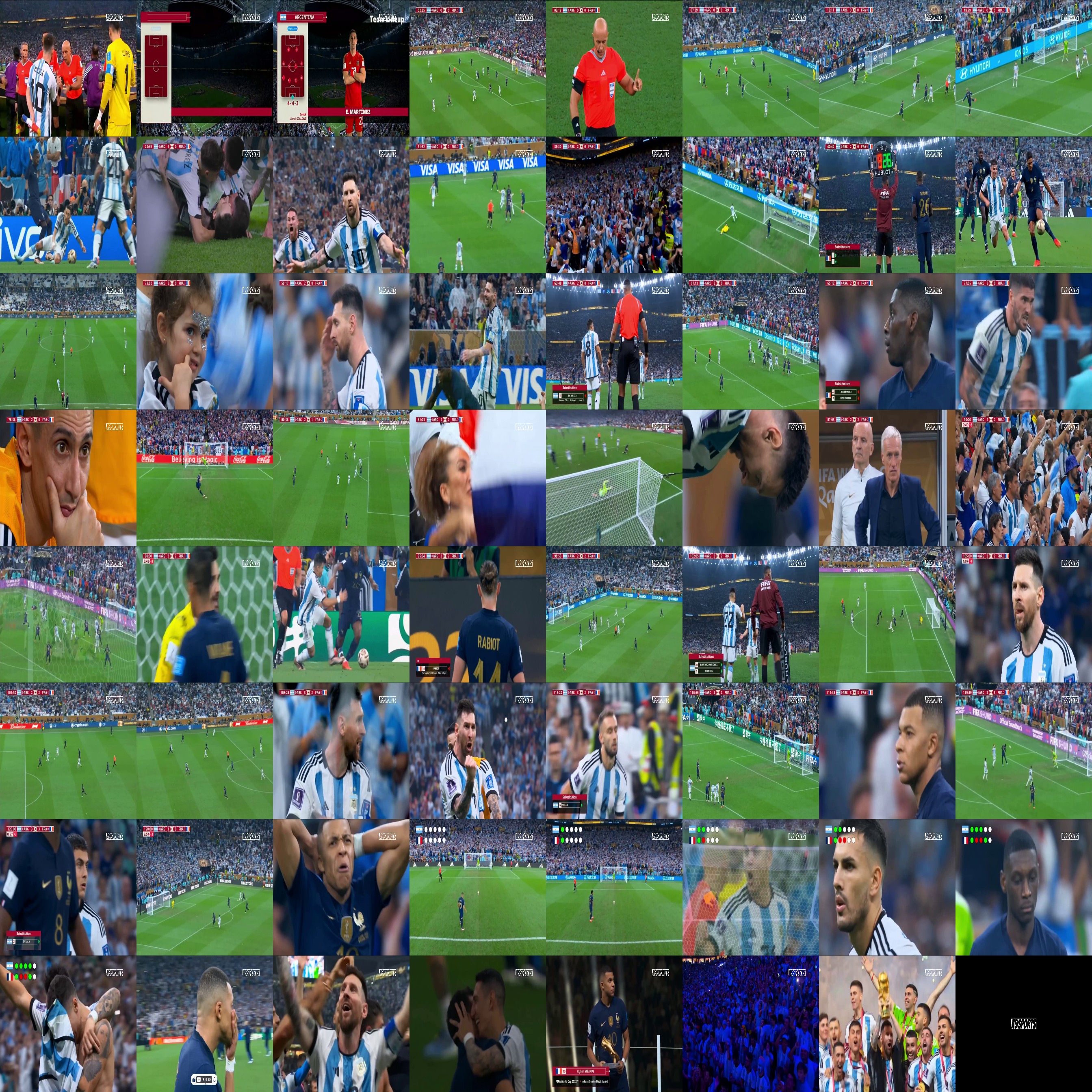}
  \\[2pt]
  {\small (a) Initial root grid (Round~0)}
  \\[8pt]
  $\boldsymbol{\Downarrow}$
  \\[8pt]
  \includegraphics[width=0.9\linewidth]{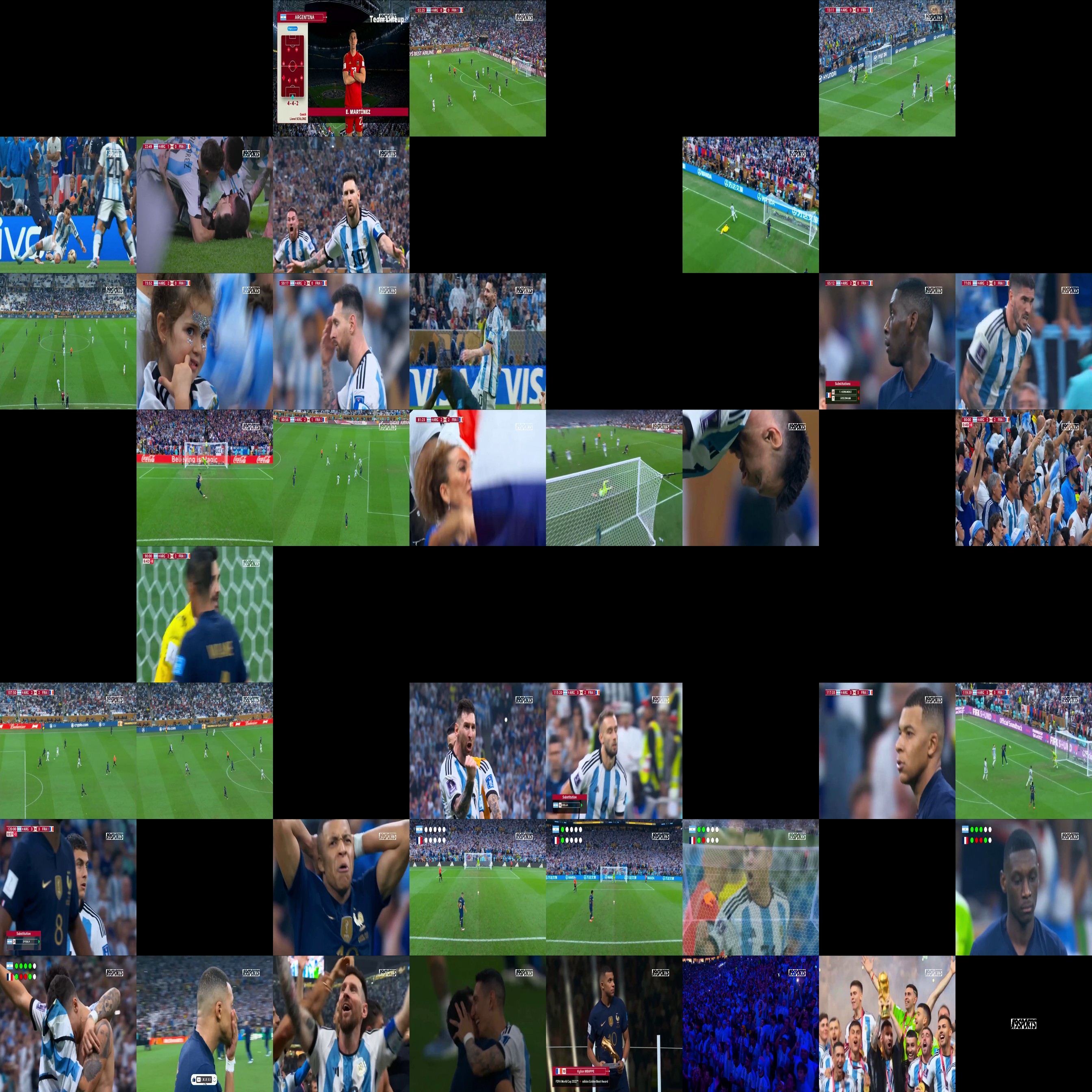}
  \\[2pt]
  {\small (b) After 8 rounds of exploration}
  \caption{\textbf{Navigation grid before and after exploration.} \textbf{(a)}~The Master's initial 8$\times$8 root grid provides a temporal overview of the full 25-minute video; each cell covers $\sim$23\,s. \textbf{(b)}~After 8 DFS rounds, explored regions are blacked out (24 of 64 cells), visually showing the coverage pattern. The system explored 4.7\% of total frames.}
  \label{fig:trace_grid}
\end{figure}

\begin{figure}[]
  \centering
  \includegraphics[width=0.99\linewidth]{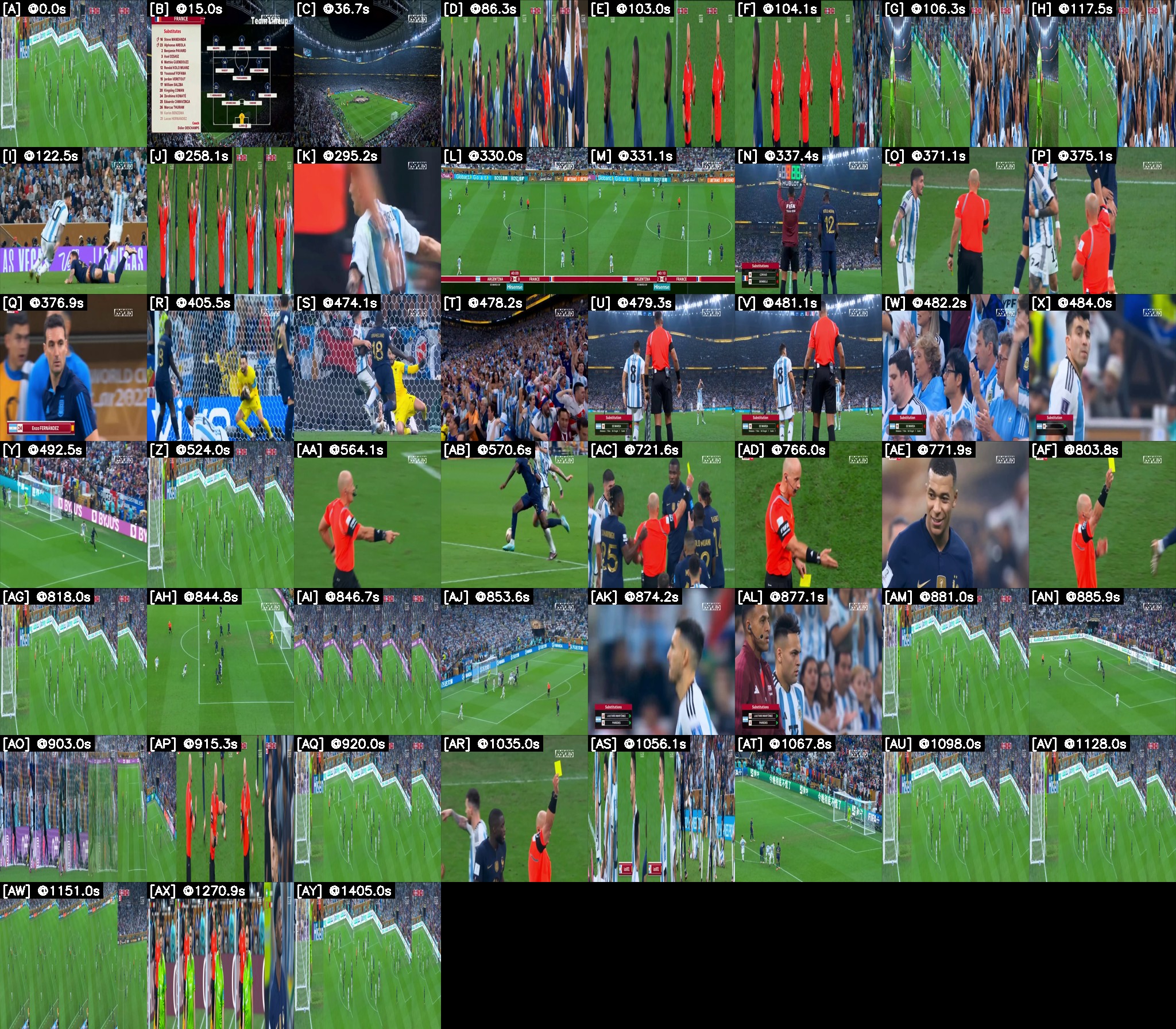}
  \caption{\textbf{Final evidence scratchpad.}
  The system's lossless visual memory after 8 rounds of exploration: 51 collected frames with burned-in labels [A]-[AY], each paired with a timestamp and natural-language description.
  Representative entries include
  \textit{``[D] @86.3\,s: Referee shows yellow card to Marcus Thuram (Minute~18),''}
  \textit{``[O] @371.1\,s: Referee gives yellow card to Enzo Fern\'{a}ndez (39$'$),''}
  Not all items are yellow-card events: the scratchpad also captures contextual evidence such as match score overlays, player close-ups, and celebration scenes, enabling the Master to cross-reference events against the full match timeline.
  This grid image is passed directly to the Master for the final decision.}
  \label{fig:trace_scratchpad}
\end{figure}

\begin{figure*}[]
  \centering
  \includegraphics[width=0.75\linewidth]{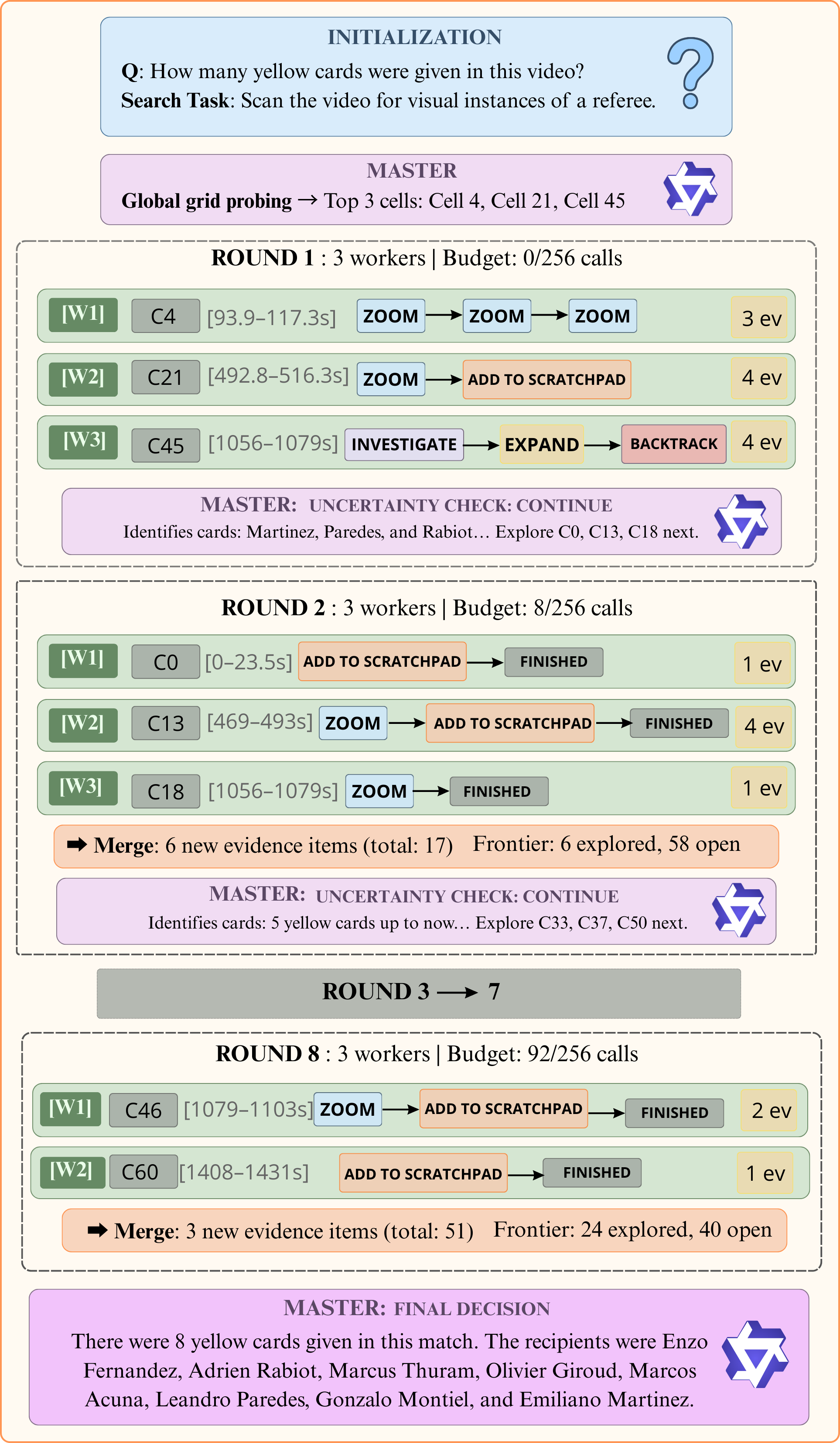}
  \caption{\textbf{Exploration trace.} Condensed log of the 8-round DFS exploration. The Master probes the root grid and dispatches 3 parallel Workers (W) per round to the highest-scored Cells (C). Workers use \textsc{Zoom}, \textsc{Investigate}, \textsc{Expand}, and \textsc{Backtrack} to drill into their assigned regions and collect evidence via \textsc{Add\_to\_Scratchpad}. After each round, the Master runs an uncertainty analysis to decide whether to continue or declare sufficiency. Evidence grows from 0 to 51 items across 8 rounds (97 VLM calls, 449K tokens), with the Master declaring \textsc{Final\_Decision} after identifying 8 distinct yellow card events, which is the correct answer.}
  \label{fig:trace_log}
\end{figure*}

We present a complete end-to-end trace of Video-RLM answering ``How many yellow cards were given in this video?'' on a 25-minute FIFA World Cup Final highlight reel (90,117 frames).
\Cref{fig:trace_grid,fig:trace_log,fig:trace_scratchpad} show the three stages of the pipeline.

% ── (1) Navigation Grid: Before → After ──

% ── (2) Exploration Trace ──

% \newpage
% ================================================================
% A. ERROR ANALYSIS
% ================================================================
\section{Detailed Error Analysis}
\label{app:error_analysis}

We extend the error analysis from \cref{sec:experiments} with a systematic characterization of failure modes.
To isolate environment failures from backbone limitations, we analyze all questions where our two backbone configurations (Qwen3.5-35B-A3B, 3B active; Gemini-3-Flash) \emph{disagree} under identical VideoAtlas configurations.
Across LongVideoBench-Long and VideoMME-Long combined, we observe 522 disagreement cases; the stronger backbone is correct in 423 of these (81\%), confirming that the dominant failure modes are backbone-dependent rather than architectural.
We identify three systematic patterns from these cases, described below.
These patterns are not mutually exclusive: a single failure may exhibit more than one.

\subsection{VLM Perception Errors}

The agent navigates to the correct temporal region but misperceives the visual content.
Two sub-patterns emerge.

\textit{(i)~Attribute confusion.}
The agent correctly identifies the scene and entities but misreads fine-grained attributes like colors, materials, spatial relationships, or on-screen text.
This is especially common when the question hinges on a single distinguishing visual feature (e.g., the color of a specific object, the label on a chart axis).

\textit{(ii)~Cross-frame inconsistency.}
The backbone produces contradictory descriptions of the same scene across different frames, then arbitrarily selects one rather than reconciling the conflict.
For example, describing the same object as ``purple/pink'' in one frame, ``white against blue'' in another, and ``blue'' in a third.

\subsection{Surface-Text Latching}

A reasoning failure where the agent anchors on a phrase in the evidence/subtitle that superficially matches a candidate answer, without verifying whether the match is contextually correct.
The agent's reasoning frequently contains high-confidence language (``the evidence \emph{explicitly states}\ldots,'' ``this \emph{directly supports} candidate X'') but that confidence is built on literal pattern-matching rather than understanding.
This is particularly problematic in documentary and educational videos, where narrators use rhetorical phrasings that contain candidate-answer keywords without implying them as the correct answer.

\subsection{Early Evidence Anchoring}

The agent commits to an answer based on the first plausible evidence item it encounters, failing to integrate later evidence that would contradict or refine its conclusion.
This failure mode interacts with the system's sufficiency mechanism: the Master may declare evidence ``sufficient'' after a single supporting item, rather than verifying coverage across all candidate answers.

% \newpage

\subsection{Impact of Backbone Quality}
\label{app:backbone_impact}

\begin{figure*}[]
  \centering
  % ── Example 1: Perception Error ──
  \begin{minipage}[t]{\linewidth}
    \centering
    \includegraphics[width=0.8\linewidth]{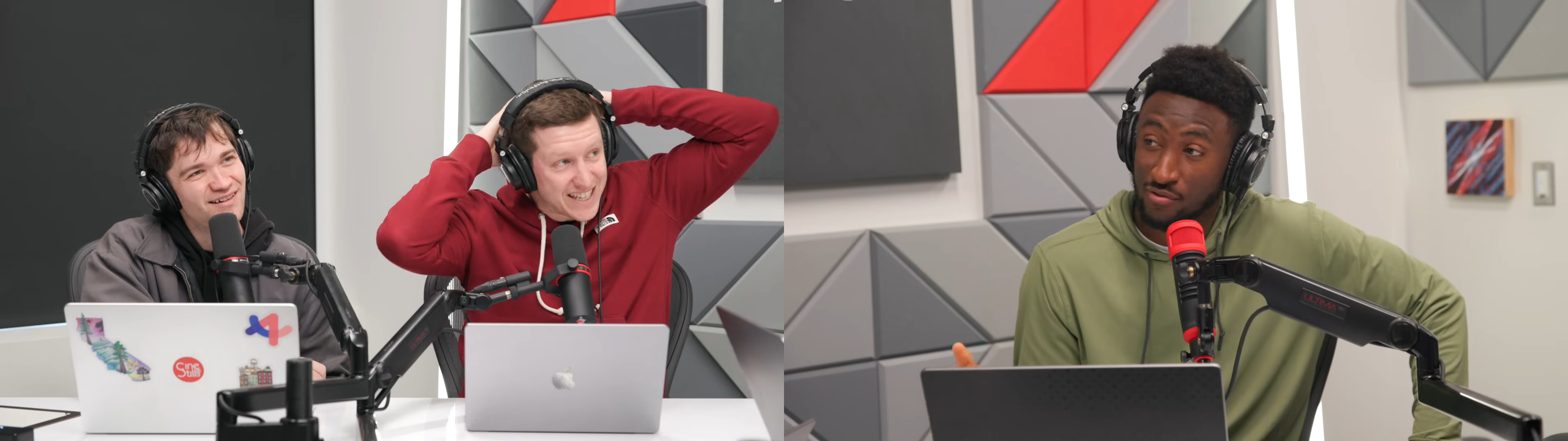}
    \\[4pt]
    \small
    \textbf{Failure mode:} VLM Perception Error \quad
    \textbf{Question:} \textit{``Which direction is the narrator in red facing relative to the narrator in green?''} \quad
    \textbf{Correct answer:} Right front (A) \\[4pt]
    \begin{tabular}{@{}p{0.38\linewidth}|p{0.38\linewidth}@{}}
      \textbf{Lightweight (3B)} & \textbf{Strong Backbone} \\
      \midrule
      ``No narrator in red appears in any frame\ldots\ neither is wearing red or green.'' &
      Identified both hosts by name from on-screen graphics across 5 evidence items. \\[3pt]
      Concluded entity does not exist. &
      Reasoned through spatial geometry of desk layout. \\[3pt]
      \xmark~\textit{Impossible to determine} & \cmark~\textit{Right front}
    \end{tabular}
  \end{minipage}

  \vspace{6pt}
  \hrule
  \vspace{6pt}

  % ── Example 2: Surface-Text Latching ──
  \begin{minipage}[t]{\linewidth}
    \centering
    \includegraphics[width=0.8\linewidth]{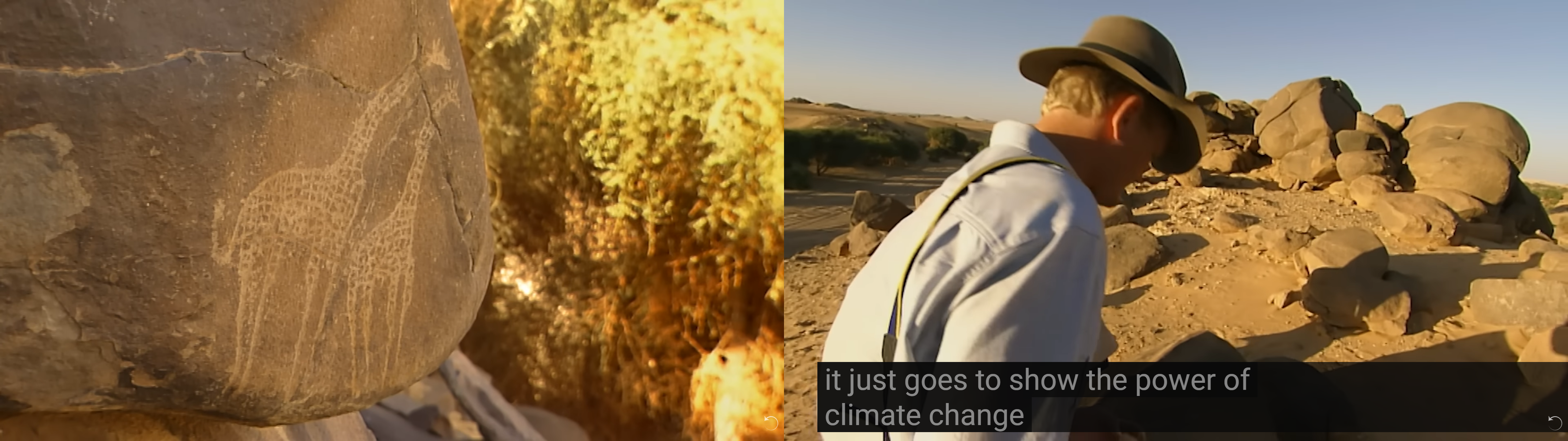}
    \\[4pt]
    \small
    \textbf{Failure mode:} Surface-Text Latching \quad
    \textbf{Question:} \textit{``What is the primary reason for the appearance of giraffe images on the rocks?''} \quad
    \textbf{Correct answer:} Climate change (C) \\[4pt]
    \begin{tabular}{@{}p{0.38\linewidth}|p{0.38\linewidth}@{}}
      \textbf{Lightweight (3B)} & \textbf{Strong Backbone} \\
      \midrule
      ``Items [A], [J], [L] \emph{explicitly state} the narration attributes giraffe images to `the creativity of the ancients.'\,'' &
      Synthesized items [M]--[O] showing whale skeletons and large animal remains in desert dunes. \\[3pt]
      Matched rhetorical phrase to option~B. &
      Linked all evidence to a dramatic past climate shift. \\[3pt]
      \xmark~\textit{Creativity of the ancients} & \cmark~\textit{Climate change}
    \end{tabular}
  \end{minipage}
   \caption{\textbf{Representative backbone failure cases.}
  \textbf{Top:}~Perception error: the lightweight backbone cannot perceive a clearly visible host wearing red and concludes the entity does not exist; the stronger backbone identifies both hosts and reasons about their spatial relationship.
  \textbf{Bottom:}~Surface-text latching: the lightweight backbone matches a rhetorical narrator phrase to a candidate answer with high confidence; the stronger backbone synthesizes broader evidence to identify the underlying scientific explanation.
  Both runs use identical VideoAtlas configurations.}
  \label{fig:error_perception}
  \label{fig:error_latching}
\end{figure*}

All three failure patterns are \emph{backbone-dependent}: switching to a stronger VLM under identical VideoAtlas configuration, prompts, and exploration budget resolves the majority of these errors without any architectural changes.
The 4:1 to 5:1 win ratio across both benchmarks reflects this consistently.
\Cref{fig:error_perception,fig:error_latching} illustrate two representative cases.

These results reinforce the main paper's finding that VideoAtlas performance scales directly with backbone capability, and suggest that the framework is well-positioned to benefit from future advances in VLM quality.

% ================================================================
% B. PER-CATEGORY ACCURACY BREAKDOWN
% ================================================================
\section{Per-Category Accuracy Breakdown}
\label{app:per_category}

LongVideoBench annotates questions along three axes: \emph{question type}, \emph{reasoning level}, and \emph{topic}.
\cref{tab:per_category_type,tab:per_category_level,tab:per_category_topic} report Video-RLM (Qwen3.5, 3B active) accuracy on both LVB-Long and LVB-10hr.

Several patterns emerge:
(1)~\emph{Sequence-type questions are hardest}: SSS (Scene Sequence Summary, 21.4\%) and SAA (33.3\%) require ordering multiple events across the full video, demanding both broad coverage and temporal precision.
(2)~\emph{Perception degrades more than relation at 10 hours}: L1-Perception drops 9.6 points (59.4$\to$49.8) vs.\ L2-Relation dropping 2.7 points (47.6$\to$44.9), suggesting that the additional temporal distance primarily harms low-level visual recognition rather than relational reasoning.
(3)~\emph{Life-Vlogs are consistently hardest}: at 35.3\% (1hr) and 26.7\% (10hr), these videos feature rapid visual changes, informal framing, and minimal subtitles, stressing the VLM's perception most severely.

\begin{table}[]
  \caption{Accuracy (\%) by question type. Categories present only in one split are marked with --.}
  \label{tab:per_category_type}
  \centering
  \small
  \setlength{\tabcolsep}{4pt}
  \begin{tabular}{@{}lcclcc@{}}
    \toprule
    & \multicolumn{2}{c}{\textbf{LVB-Long}} & \phantom{a} & \multicolumn{2}{c}{\textbf{LVB-10hr}} \\
    \cmidrule(lr){2-3}\cmidrule(lr){5-6}
    Question Type & $n$ & Acc & & $n$ & Acc \\
    \midrule
    S2A  & 30 & 66.7 && 27 & 63.0 \\
    T2E  & 33 & 63.6 && 33 & 60.6 \\
    S2E  & 42 & 61.9 && 42 & 50.0 \\
    S2O  & 33 & 60.6 && 33 & 42.4 \\
    O3O  & 30 & 60.0 && -- & -- \\
    SOS  & 42 & 57.1 && -- & -- \\
    TOS  & 21 & 57.1 && -- & -- \\
    T2O  & 18 & 55.6 && 18 & 44.4 \\
    T2A  & 42 & 54.8 && 42 & 47.6 \\
    E3E  & 51 & 54.9 && 48 & 54.2 \\
    O2E  & 24 & 54.2 && -- & -- \\
    TAA  & 36 & 52.8 && 36 & 44.4 \\
    E2O  & 12 & 50.0 && 12 & 25.0 \\
    T3O  & 39 & 48.7 && 33 & 30.3 \\
    T3E  & 33 & 48.5 && 30 & 46.7 \\
    SAA  & 36 & 33.3 && -- & -- \\
    SSS  & 42 & 21.4 && -- & -- \\
    \bottomrule
  \end{tabular}
\end{table}

\begin{table}[]
  \caption{Accuracy (\%) by reasoning level.}
  \label{tab:per_category_level}
  \centering
  \small
  \begin{tabular}{@{}lcclcc@{}}
    \toprule
    & \multicolumn{2}{c}{\textbf{LVB-Long}} & \phantom{a} & \multicolumn{2}{c}{\textbf{LVB-10hr}} \\
    \cmidrule(lr){2-3}\cmidrule(lr){5-6}
    Level & $n$ & Acc & & $n$ & Acc \\
    \midrule
    L1 -- Perception & 234 & 59.4 && 207 & 49.8 \\
    L2 -- Relation   & 330 & 47.6 && 147 & 44.9 \\
    \bottomrule
  \end{tabular}
\end{table}

\begin{table}[]
  \caption{Accuracy (\%) by topic category.}
  \label{tab:per_category_topic}
  \centering
  \small
  \begin{tabular}{@{}lcclcc@{}}
    \toprule
    & \multicolumn{2}{c}{\textbf{LVB-Long}} & \phantom{a} & \multicolumn{2}{c}{\textbf{LVB-10hr}} \\
    \cmidrule(lr){2-3}\cmidrule(lr){5-6}
    Topic & $n$ & Acc & & $n$ & Acc \\
    \midrule
    Cooking-Recipes       & 42 & 64.3 && 30 & 56.7 \\
    Knowledge-Geography   & 54 & 63.0 && 45 & 46.7 \\
    Travel-Guides         & 54 & 61.1 && 30 & 53.3 \\
    Knowledge-STEM        & 75 & 54.7 && 48 & 52.1 \\
    Knowledge-History     & 69 & 53.6 && 42 & 54.8 \\
    Movie-Recaps          & 63 & 52.4 && 36 & 47.2 \\
    News-Programs         & 48 & 52.1 && 27 & 40.7 \\
    Knowledge-CS          & 51 & 47.1 && 30 & 40.0 \\
    Knowledge-Art         & 57 & 42.1 && 36 & 52.8 \\
    Life-Vlogs            & 51 & 35.3 && 30 & 26.7 \\
    \bottomrule
  \end{tabular}
\end{table}

% ── Error Examples (combined into one figure) ──

% ================================================================
% C. COMPUTE BREAKDOWN
% ================================================================
\section{Compute Breakdown}
\label{app:compute}

\cref{tab:token_breakdown} decomposes the average token consumption per question into Master and Worker contributions. The Master accounts for a small fraction of total tokens (global probing, uncertainty analysis, final decision), while the bulk of compute is spent on Worker exploration. This confirms that the parallel architecture efficiently distributes work: the Master acts as a lightweight coordinator while Workers perform the heavy visual exploration.

\begin{table}[h]
  \caption{Average per-question compute breakdown for Video-RLM (Qwen3.5, 3B active).}
  \label{tab:token_breakdown}
  \centering
  \small
  \setlength{\tabcolsep}{3pt}
  \begin{tabular}{@{}l rrr ccc@{}}
    \toprule
    & \multicolumn{3}{c}{Avg. Tokens} & Suff. & Expl. & Evid. \\
    \cmidrule(lr){2-4}
    Benchmark & Master & Worker & Total & Checks & Rounds & Items \\
    \midrule
    LVB-Long  & 28K & 121K & 149K & 1.8 & 2.0 & 6.2 \\
    LVB-10hr  & 30K & 219K & 250K & 1.9 & 2.3 & 7.2 \\
    \bottomrule
  \end{tabular}
\end{table}

At 10 hours, Worker tokens increase by 80\% (121K$\to$219K) while Master tokens increase by only 10\% (27.5K$\to$30.4K), confirming that the Master's coordination overhead scales minimally with video duration. The additional Worker cost reflects deeper exploration (2.3 vs.\ 2.0 rounds) needed to locate evidence in a 10$\times$ longer video, yet the increase is far sub-linear relative to the 10$\times$ duration increase, consistent with the logarithmic scaling property described in \cref{sec:log_scaling}. Evidence items increase modestly (6.2$\to$7.2), suggesting the system explores more but collects evidence at a similar density.

\subsection{Multimodal Token Efficiency}
\label{app:mm_cache}

During DFS exploration, each worker re-examines the same grid view across multiple reasoning steps (e.g., \textsc{Add\_to\_Scratchpad} steps that do not change the navigation state), creating inherent redundancy in visual token processing.
For \textbf{Qwen (self-hosted via vLLM)}, this is handled transparently: vLLM's automatic multimodal prefix caching detects repeated image token prefixes across requests and serves KV cache hits without any code changes.
\cref{tab:cache_rates} reports the measured hit rates across video durations.

\begin{table}[h]
  \caption{vLLM multimodal prefix cache hit rates for Qwen3.5 across video durations.}
  \label{tab:cache_rates}
  \centering
  \small
  \begin{tabular}{@{}l cc@{}}
    \toprule
    Duration & Avg.\ Hit Rate (\%) & Std.\ (\%) \\
    \midrule
    \multicolumn{3}{@{}l}{\textit{LongVideoBench}} \\
    1\,min   & 60.9 & 7.0 \\
    10\,min  & 48.2 & 7.1 \\
    1\,hr    & 42.7 & 7.5 \\
    3\,hr    & 50.7 & 12.4 \\
    5\,hr    & 48.9 & 11.2 \\
    10\,hr   & 41.6 & 8.8 \\
    \midrule
    \multicolumn{3}{@{}l}{\textit{Video-MME (no subs)}} \\
    10\,hr   & 35.5 & 8.9 \\
    \bottomrule
  \end{tabular}
\end{table}

% ================================================================
% D. PROMPT TEMPLATES
% ================================================================
\section{Prompt Templates}
\label{app:prompts}

We include the complete prompt templates used by VideoAtlas.
All prompts are zero-shot (no in-context examples).
The same templates are used across all benchmarks and video durations without modification.

\subsection{Search Task Extraction}
\label{app:prompt_search_task}
A text-only call that converts the raw query and answer candidates into a concrete visual search task.

\begin{small}
\begin{verbatim}
Convert this question + choices into 
a concrete SEARCH TASK for exploring a video.

Question: "{query}"
Choices:
{candidates}

Describe EXACTLY what to look for visually.
Be specific about scenes, objects, 
text overlays, or transitions 
that would confirm each choice.
Output only the search task, no preamble.
\end{verbatim}
\end{small}

\subsection{Master: Global Probing}
\label{app:prompt_probe}
The Master examines the root grid (with dead zones blacked out) and ranks the top-$N$ cells for worker assignment.

\begin{small}
\begin{verbatim}
You are analyzing a KxK grid of frames sampled
from a SINGLE video in chronological order
(left-to-right, top-to-bottom).

**QUERY:** "{query}"

**GRID CELLS:**
{context_str}

Pick EXACTLY {top_n} cells (no more, no fewer)
most likely to help answer the query.

**OUTPUT (raw JSON, EXACTLY {top_n} entries):**
{"top": [{"id": <cell_id>}, ...]}
\end{verbatim}
\end{small}

\subsection{Master: Uncertainty Analysis}
\label{app:prompt_uncertainty}
After each round with new evidence, the Master performs three tasks in one call: sufficiency check, explore suggestions, and noise erasure.

\begin{small}
\begin{verbatim}
**UNCERTAINTY ANALYSIS**

You are the MASTER coordinator analyzing 
search progress for a video question.

**QUERY:** "{query}"
**ANSWER CHOICES:**
{candidates}
**EVIDENCE COLLECTED SO FAR:**
{evidence_text}
**EXPLORATION PROGRESS:**
{progress_text}
**NAVIGATION GRID (blacked-out = explored):**
{context_str}

**YOUR 3 TASKS (do all in one response):**

1. **UNCERTAINTY CHECK:** For each answer choice,
   do you have sufficient evidence?
2. **EXPLORE SUGGESTIONS:** Suggest up to {N}
   regions. ONLY suggest non-blacked-out cells.
   - Cell IDs from the grid
   - Custom time ranges {"start","end"} (<60s)
3. **ERASE NOISE:** ONLY erase evidence completely
   unrelated to query, task, and ALL choices.
   Keep partial evidence. When in doubt, keep it.

**If sufficient:** {"action": "FINAL_DECISION",...}
**Otherwise:** {"action": "CONTINUE",
  "reasoning": "...", "explore": [...],
  "erase": [...]}
\end{verbatim}
\end{small}

\subsection{Worker: Exploration Step}
\label{app:prompt_worker}
Each worker receives a grid view of its assigned region and the available tool set.
The prompt includes a 1-sentence summary of the previous step's outcome (conversation history).

\begin{small}
\begin{verbatim}
SEARCH TASK: "{search_task}"
QUERY: "{query}"

You are exploring [{start}-{end}s]
({pct}% through a {duration}s video, depth {d}).
Grid: {K}x{K}, chronological L-to-R, T-to-B.

**CELLS:**
{context_str}

**PREVIOUS:** {prev_summary}

**RULES:**
- EXPAND into promising cells to zoom in
- Use ZOOM only when you found a relevant scene
  and need a closer high-resolution look
- Use INVESTIGATE only when you found the anchor
  scene and need to check what happens before/after
- ADD_TO_SCRATCHPAD with timestamp, description,
  and confidence when you find evidence
- FINISHED when region has no relevant content

Pick ONE action. Be precise with timestamps.
\end{verbatim}
\end{small}

\subsection{Master: Final Decision}
\label{app:prompt_final}
After exploration terminates, the Master sees the evidence scratchpad grid and evaluates each candidate.

\begin{small}
\begin{verbatim}
You are making a FINAL DECISION based on all
collected visual evidence.

QUERY: "{query}"
ANSWER CHOICES:
{candidates}

EVIDENCE (see grid image with burned-in labels):
{evidence_descriptions}

For EACH choice: state which evidence supports
or contradicts it. Then select the best-supported
answer.

**OUTPUT:**
{"answer": <choice index>, "reasoning": "..."}
\end{verbatim}
\end{small}